\pdfoutput=1

\documentclass[11pt]{article}

\usepackage[]{acl}

\usepackage{times}
\usepackage{latexsym}

\usepackage[T1]{fontenc}

\usepackage[utf8]{inputenc}

\usepackage{microtype}

\usepackage{graphicx}
\usepackage{arydshln}
\usepackage{xcolor}
\usepackage{amsmath}

\DeclareMathOperator*{\argmin}{arg\,min}
\usepackage{amssymb}
\usepackage{subcaption}
\usepackage{caption}
\usepackage{multicol}
\usepackage{multirow}
\usepackage{booktabs}
\def\@fnsymbol#1{\ensuremath{\ifcase#1\or *\or \dagger\or \ddagger\or
   \mathsection\or \mathparagraph\or \|\or **\or \dagger\dagger
   \or \ddagger\ddagger \else\@ctrerr\fi}}
\newcommand{\ssymbol}[1]{^{\@fnsymbol{#1}}}
\usepackage{nicematrix}

\definecolor{officegreen}{rgb}{0.0, 0.5, 0.0}

\title{Revisit Few-shot Intent Classification with PLMs: \\
Direct Fine-tuning vs. Continual Pre-training  }

\author{
Haode Zhang$^1$ \quad Haowen Liang$^1$ \quad  \bf Liming Zhan$^1$ \\ \quad \bf Albert Y.S. Lam$^2$ \quad \bf Xiao-Ming Wu$^1$\Thanks{~ Corresponding author.}\\ 
Department of Computing, The Hong Kong Polytechnic University, Hong Kong S.A.R.$^1$ \\
Fano Labs, Hong Kong S.A.R.$^2$ \\
{\tt \small \{haode.zhang,michaelhw.liang,lmzhan.zhan\}@connect.polyu.hk} \\
{\tt \small xiao-ming.wu@polyu.edu.hk, albert@fano.ai} \\
}

\begin{document}
\maketitle
\begin{abstract}
We consider the task of few-shot intent detection, which involves training a deep learning model to classify utterances based on their underlying intents using only a small amount of labeled data.
The current approach to address this problem is through continual pre-training, i.e., fine-tuning pre-trained language models (PLMs) on external resources (e.g., conversational corpora, public intent detection datasets, or natural language understanding datasets) before using them as utterance encoders for training an intent classifier.
In this paper, we show that continual pre-training may not be essential, since the overfitting problem of PLMs on this task may not be as serious as expected. 
Specifically, we find that directly fine-tuning PLMs on only a handful of labeled examples already yields decent results compared to methods that employ continual pre-training, and the performance gap diminishes rapidly as the number of labeled data increases. To maximize the utilization of the limited available data, 
we propose a context augmentation method and leverage sequential self-distillation to boost performance. Comprehensive experiments on real-world benchmarks show that given only two or more labeled samples per class, direct fine-tuning outperforms many strong baselines that utilize external data sources for continual pre-training.
The code can be found at \url{https://github.com/hdzhang-code/DFTPlus}.

\end{abstract}

\section{Introduction}
Intent detection is a critical module in task-oriented dialogue systems. The target is to classify utterances according to user intents. 
Recent progress in intent detection highly relies on deep models and datasets with well-crafted annotations. Using large-scale models or datasets has been recognized as a de facto recipe for many tasks in natural language processing (NLP) including intent detection . However, large training datasets are often not available due to the cost of labeling.
Therefore, few-shot intent detection, which aims to train a classifier with only a few labeled examples, has attracted considerable attention in recent years~\cite{dopierre-etal-2021-protaugment, zhang-etal-2022-fine, mi2022cins}.

\begin{figure}[t]
    \centering
    \includegraphics[scale=0.33]{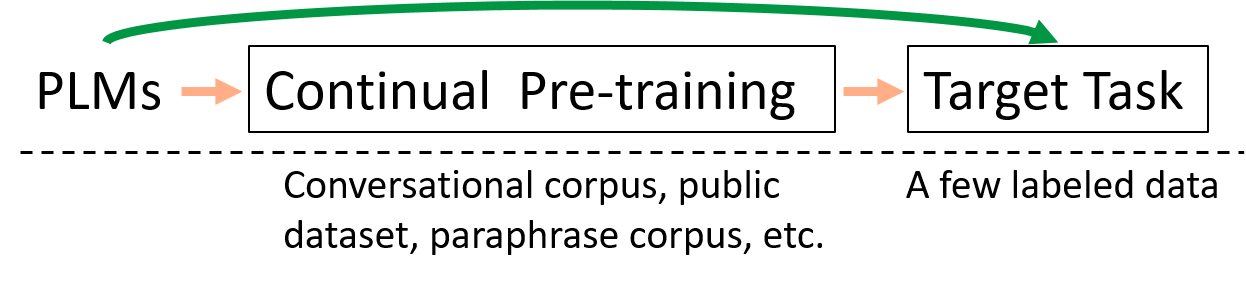}
    \caption{Illustration of continual pre-training  (\textcolor{orange}{orange})   and direct fine-tuning~(\textcolor{officegreen}{green}).}
    \label{figure: introduction_direct_tackle_target_task}
\end{figure}
\begin{figure*}[t]
\centering
    \centering
    \includegraphics[scale=0.56]{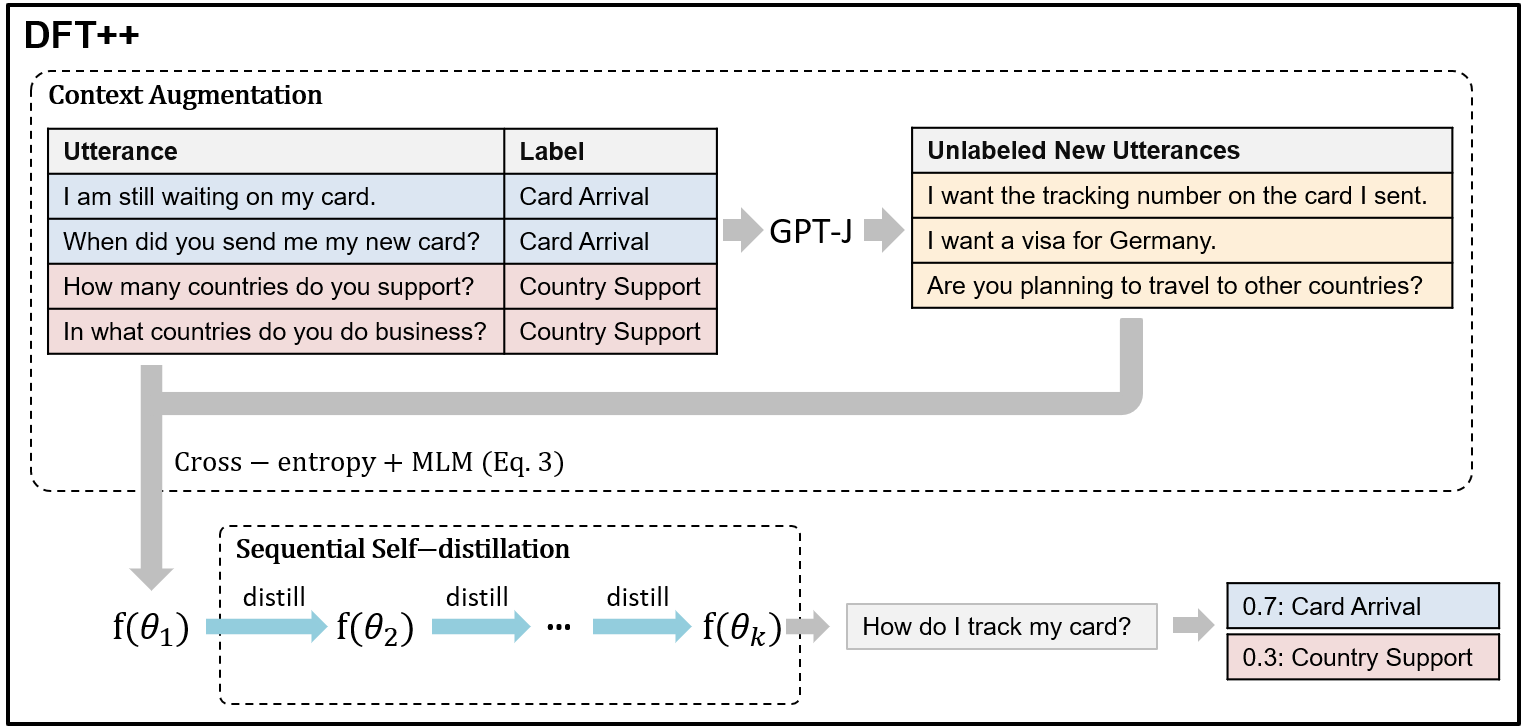}
\caption{Illustration of DFT++ with $2$ classes and $2$ labeled examples per class. GPT-J is employed to generate contextually relevant unlabeled utterances. Sequential self-distillation is performed to further boost the performance.}
\label{figure: idea_illustration}
\end{figure*}

The main obstacle for few-shot learning is commonly believed to be overfitting, i.e. the model trained with only a few examples tends to overfit to the training data and 
perform much worse on test data~\cite{vinyals2016matching, zhang-etal-2022-fine}. To alleviate the problem, the mainstream approach is to transfer knowledge from \emph{external resources} such as another labeled dataset, which has been widely used for few-shot image classification~\cite{fei2006one, snell2017prototypical} and few-shot intent detection~\cite{yu2018diverse, geng2019few, nguyen2020dynamic}.

Since recently emerged large-scale pre-trained language models (PLMs) 
have achieved great success in various NLP tasks, most recent few-shot intent detection methods propose to fine-tune PLMs on external resources before applying them on the target task, which is known as \emph{continual pre-training}~\cite{gururangan2020don, ye-etal-2021-crossfit}, as illustrated in Fig~\ref{figure: introduction_direct_tackle_target_task}. The external resources utilized for continual pre-training include conversational corpus~\cite{wu2020tod, mehri2020dialoglue, vulic-etal-2021-convfit}, natural language understanding datasets~\cite{zhang2020discriminative}, public intent detection datasets~\cite{zhang-etal-2021-effectiveness-pre, yu2021few}, and paraphrase corpus~\cite{ma-etal-2022-effectiveness}. 
While these methods have achieved state-of-the-art results, the use of external training corpora induces extra data processing effort (e.g., SBERT-Paraphrase~\citet{ma-etal-2022-effectiveness} uses 83 million sentence pairs from 12 datasets) 
as well as model bias (e.g., the trained model may be biased to the intent classes used in continual pre-training)~\cite{xia2020cg, xia2021pseudo, nguyen2020dynamic}.

It is commonly believed that directly fine-tuning PLMs with a small amount of data may generate unacceptable variance~\cite{lee2020mixout, dodge2020fine}. However, it has been recently found that the instability may be caused by incorrect use of optimizer and insufficient training~\cite{mosbach2021stability, zhang2020revisiting}. Further, some studies~\cite{hao2019visualizing, li2019exploiting} have revealed that in sentiment analysis and paraphase detection tasks, when directly fine-tuned with a small dataset, PLMs sush as BERT~\cite{devlin2018bert} demonstrate a certain level of resilience to overfitting.
Therefore, a thorough investigation is needed to explore the direct fine-tuning of PLMs for few-shot intent detection.
In this work, we make the following contributions: 
\begin{itemize}
    \item  We take an empirical investigation into the overfitting issue when directly fine-tuning PLMs on few-shot intent detection tasks. Our study suggests that overfitting may not be a significant concern, since the test performance improves rapidly as the size of training data increases. Further, the model's performance does not degrade  as training continues. It implies that early stopping is not necessary, which is often employed to prevent overfitting in few-shot learning and requires an additional set of labeled data for validation.
    
    

    
     \item  We find that direct fine-tuning (DFT) already yields decent results compared with continual pre-training methods. We further devise a DFT++ framework to fully exploit the given few labeled data and boost the performance. DFT++ introduces a novel \emph{context augmentation} mechanism by using a generative PLM to generate \emph{contextually relevant unlabeled data} to enable better adaptation to target data distribution, as well as a sequential self-distillation mechanism to exploit the multi-view structure in data. A comprehensive evaluation shows that DFT++  outperforms state-of-the-art continual pre-training methods with only the few labeled data provided for the task, without resorting to external training corpora. 
\end{itemize}

\section{Direct Fine-tuning}
We investigate a straightforward approach for few-shot intent detection -- directly fine-tuning (DFT) PLMs with the few-shot data at hand. 
However, it is a common belief that such a process may lead to severe overfitting. Before 
going into detail,
we first formally define the problem.

\subsection{Problem Formulation}
\label{section: Fine-tuning PLMs}
Few-shot intent detection aims to train an intent classifier with only a small labeled dataset $\mathcal{D}=\{(x_i,y_i)\}_{N}$, where $N$ is the dataset size, $x_i$ denotes the $i_\text{th}$ utterance, and $y_i$ is the label. The number of samples per label is typically less than $10$.

We follow the standard practice~\cite{sun2019fine, zhang-etal-2021-effectiveness-pre} to apply a linear classifier on top of the utterance representations:
\begin{equation}
    \begin{split}
        \text{p}(y|\mathbf{h}_i) = \text{softmax}\left(\mathbf{W}\mathbf{h}_i + \mathbf{b}\right) \in \mathbb{R}^L,
    \end{split}
    \label{equation: linear layer classifier}
\end{equation}
where $\mathbf{h}_i \in \mathbb{R}^{d}$ is the representation of the $i_\text{th}$ utterance in $\mathcal{D}$, $\mathbf{W} \in \mathbb{R}^{L \times d}$ and $\mathbf{b} \in \mathbb{R}^{L}$ are the parameters of the linear layer, and $L$ is the number of classes. We use the representation of the [CLS] token as the utterance embedding $\mathbf{h}_i$. The model parameters $\theta=\left\{\phi, \mathbf{W}, \mathbf{b}\right\}$, with $\phi$ being the parameters of the PLM, are trained on $\mathcal{D}$. We use a cross-entropy loss $\mathcal{L}_{\text{ce}}\left(\cdot\right)$ to learn the model parameters:
\begin{equation}
        \theta = \argmin_{\theta} \mathcal{L}_{\text{ce}}\left(\mathcal{D};\theta\right).
    \label{equation: arg min cross entropy}
\end{equation}

Unlike the popular approach of continual pre-training~\cite{zhang2020discriminative, zhang-etal-2022-fine, zhang-etal-2021-shot}, DFT fine-tunes PLMs directly on the few-shot data, which may experience overfitting, leading to suboptimal performance. To examine this issue, we conduct the following experiments.

\subsection{Experiments}\label{sec:DFT-experiments}
\paragraph{Datasets} We utilize four  large-scale practical datasets.
\textbf{HINT3}~\cite{arora2020hint3} is created from live chatbots with 51 intents. 
\textbf{BANKING77}~\cite{casanueva2020efficient} is a fine-trained dataset focusing on banking services, containing 77 intents.
\textbf{MCID} ~\cite{arora2020cross} is a cross-lingual dataset for ``Covid-19'' with 16 intents, and we use the English version only.
\textbf{HWU64}~\cite{DBLP:conf/iwsds/LiuESR19} is a large-scale multi-domain dataset with 64 intents. The statistics of the datasets are given in Table~\ref{table: Dataset statistics}. To simulate few-shot scenarios, we randomly sample $K$ samples per label from the training set of each dataset to form the dataset $\mathcal{D}$. 
\begin{table}[h]
\centering
\small
\begin{tabular}{lcccc}
\toprule
Dataset   &  \#Intent & \#Train & \#Dev & \#Test \\
\midrule
OOS       &  150   & 15000 & 3000 & 4500    \\
BANKING77 &  77    & 10003  & 0 & 3080    \\
HINT3     &  51    & 1579  & 0 & 676        \\  
HWU64     &  64    & 8954  & 1076 & 1076    \\
MCID      &  16    & 1258 & 148 & 339     \\
\bottomrule
\end{tabular}
\caption{Dataset statistics.}
\label{table: Dataset statistics}
\end{table}


\paragraph{Baselines}
To evaluate DFT, we compare it against IsoIntentBERT~\cite{zhang-etal-2022-fine}, a competitive baseline applying continual pre-training with public intent detection datasets. We follow the original work to pre-train BERT on OOS~\cite{larson2019evaluation}, a multi-domain public intent detection dataset containing diverse semantics, and then perform in-task fine-tuning on the small dataset $\mathcal{D}$.



\begin{figure}[h]
\centering
    \begin{subfigure}[t]{0.23\textwidth}
        \includegraphics[scale=0.24]{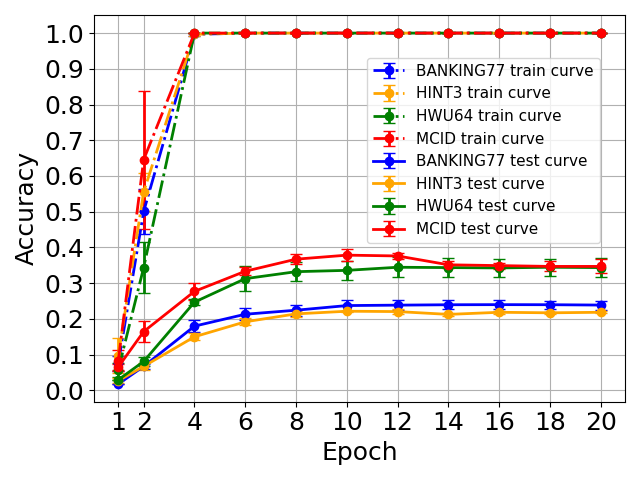}
        \caption{BERT, 1-shot.}
        \label{figure: pilot_study_shot1, bert}
    \end{subfigure}
    \begin{subfigure}[t]{0.23\textwidth}
        \includegraphics[scale=0.24]{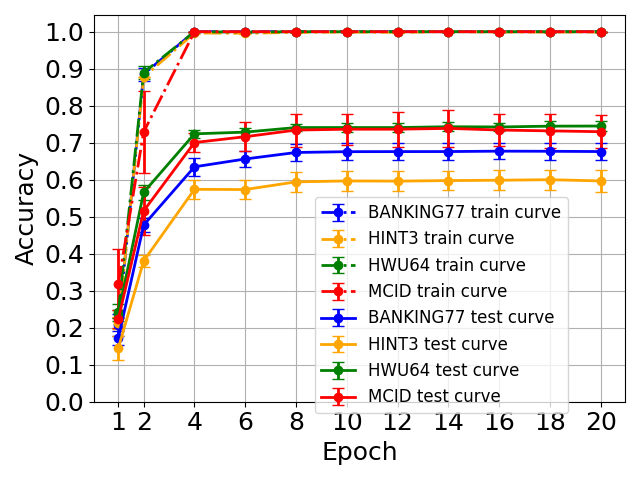}
        \caption{BERT, 5-shot.}
        \label{figure: pilot_study_shot5, bert}
    \end{subfigure}    
    \begin{subfigure}[t]{0.23\textwidth}
        \includegraphics[scale=0.24]{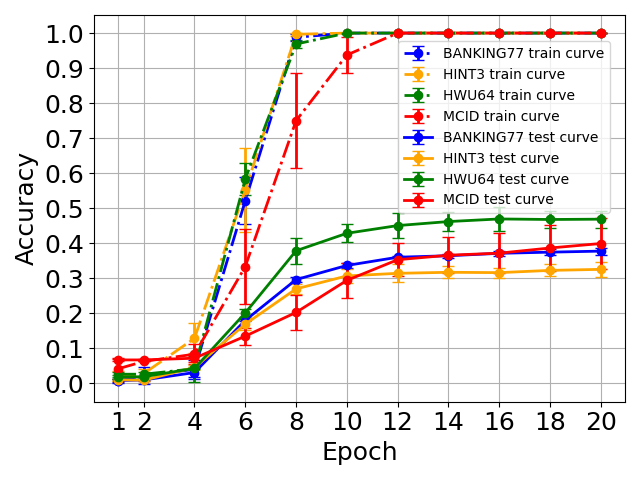}
        \caption{RoBERTa, 1-shot.}
        \label{figure: pilot_study_shot1, roberta}
    \end{subfigure}
    \begin{subfigure}[t]{0.23\textwidth}
        \includegraphics[scale=0.24]{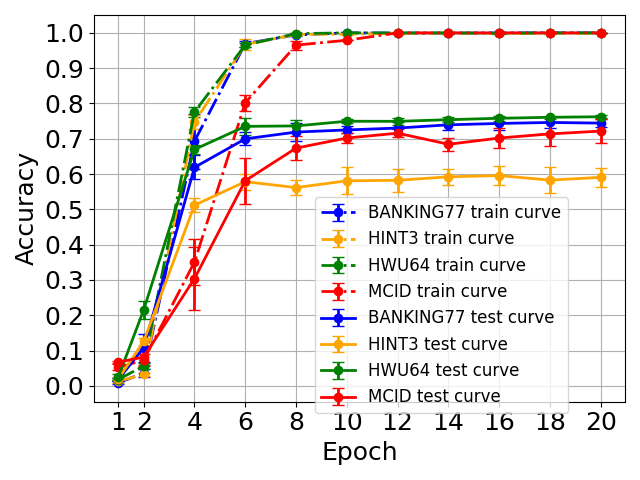}
        \caption{RoBERTa, 5-shot.}
        \label{figure: pilot_study_shot5, roberta}
    \end{subfigure}    
\caption{Training and test learning curves of DFT with BERT and RoBERTa as text encoder respectively.}
\label{figure: overfit_analysis}
\end{figure}
\begin{figure}[ht]
    \centering
    \includegraphics[scale=0.40]{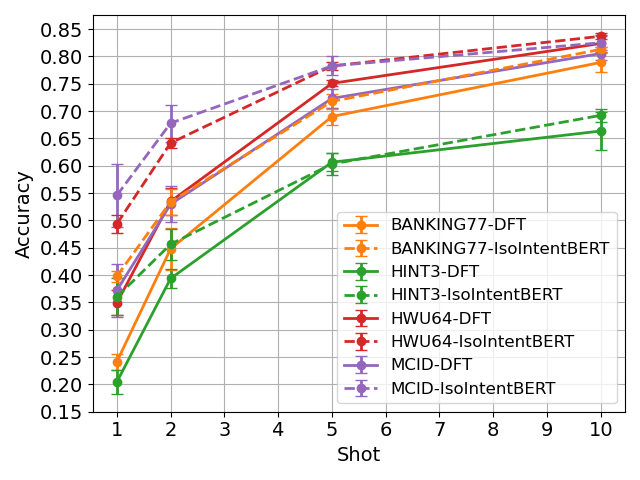}
    \caption{Comparison between DFT (solid lines) and IsoIntentBERT (dashed lines). The benefit from continued pre-training(IsoIntentBERT) decays quickly.}
    \label{figure: pilot_study_with_and_no_pretraining}
\end{figure}

\paragraph{Results and Findings} We plot the learning curves of DFT in Fig.~\ref{figure: overfit_analysis}, where the following observations can be drawn. First, comparing the results in 1-shot and 5-shot scenarios, the test performance of DFT improves drastically as the number of labeled examples rises from 1 to 5, leading to a fast reduction in the performance gap between the training and test performance. Second, the test performance does not deteriorate as the training progresses, and the learning curves exhibit a flat trend. These observations are consistent across various datasets and different models (BERT and RoBERTa), including both 1-shot and 5-shot\footnote{We observe the same patterns when further increasing the shot number beyond 5.} scenarios.
The observations also align with previous findings in sentiment analysis~\cite{li2019exploiting} and paraphrase detection~\cite{hao2019visualizing} tasks. 


The flat learning curves indicate that early stopping is not necessary, which is often used to prevent overfitting and requires an additional set of labeled data. This is important for practitioners because model selection has been identified as a roadblock for \emph{true few-shot learning}~\cite{perez2021true}, where the labeled data is so limited that it is not worth setting aside a portion of it for early stopping. On the other hand, the rapidly reduced performance gap between DFT and IsoIntentBERT (Fig.~\ref{figure: pilot_study_with_and_no_pretraining}) casts doubt on the necessity of continual pre-training. 
Thus, we raise an intriguing question:
\begin{itemize}
    \item With only the given few labeled data, is it possible to achieve comparable or better performance than continual pre-training methods?
\end{itemize}
Our attempt to answer the question leads to DFT++, a framework designed to fully exploit the given few labeled data, which provides an affirmative answer.

\section{Push the Limit of Direct Fine-Tuning}
To push the limit of few-shot intent detection with only a few labeled data at hand and without using any external training corpora,
DFT++ introduces two mechanisms, as shown in Fig.~\ref{figure: idea_illustration}. The first is a novel context augmentation mechanism, wherein the few data are used to prompt a generative PLM to generate contextually relevant unlabeled utterances to better model target data distribution. The second is a sequential self-distillation mechanism. 

\subsection{Context Augmentation}
\begin{figure}[ht]
    \centering
    \includegraphics[scale=0.45]{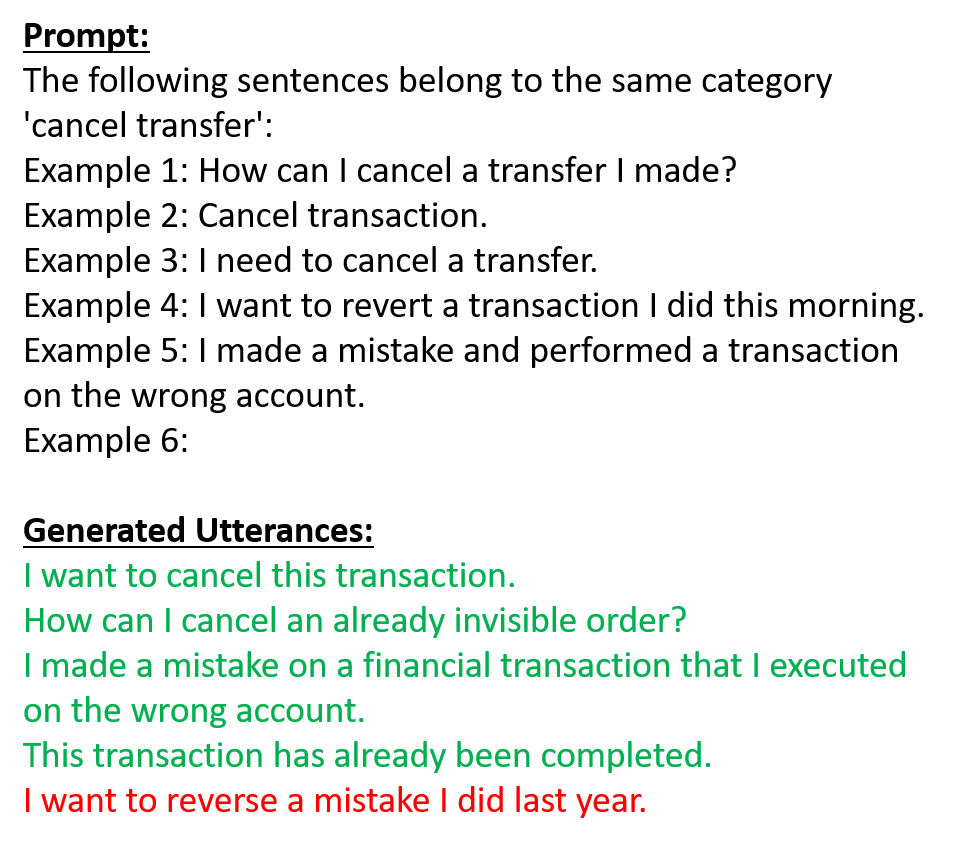}
    \caption{An example of the prompt and generated utterances in a $5$-shot scenario. \textcolor{officegreen}{Green} utterances are successful cases, while the \textcolor{red}{red} one is a failure case.}
    \label{figure: latex_figures_pilot_study_prompt_example}
\end{figure}
Unlike continual pre-training methods that leverage external training corpora, we use the few data to solicit knowledge from generative PLMs. An intuitive way is data augmentation, which prompts the model to generate new utterances with the given intent class. However, as suggested  by~\citet{sahu-etal-2022-data} and our analysis (Section~\ref{subsection: analysis}), data augmentation for intent detection with tens of intent classes is challenging. Hence, we propose to exploit contextual relevance in an unsupervised manner instead.
Specifically, for each intent class, we compose the few data into a prompt and then feed it to GPT-J~\cite{gpt-j}, a powerful generative PLM, to generate novel unlabeled utterances. Fig.~\ref{figure: latex_figures_pilot_study_prompt_example} gives an example of the prompt and generated results. The generated unlabeled data is combined with the given utterances in $\mathcal{D}$ to compose a corpus $\mathcal{D}_\text{aug}=\{x_i\}_i$, which can be used for masked language modeling (MLM). Hence, the model parameters $\theta$ are learned by simultaneously minimizing both the cross-entropy loss $\mathcal{L}_\text{ce}$ and the MLM loss $\mathcal{L}_{\text{mlm}}$:
\begin{equation}
    \theta = \argmin_{\theta}\left(    \mathcal{L}_\text{ce}(\mathcal{D};\theta)+\lambda \mathcal{L}_{\text{mlm}}(\mathcal{D}_{\text{aug}};\theta)
    \right), 
    \label{equation: loss, ce and mlm}
\end{equation}
where $\lambda$ is a balancing parameter.

Notice that there is a critical difference between the proposed context augmentation and conventional data augmentation methods. Context augmentation generates contextually relevant data (i.e., utterances with similar context to the given input but not necessarily belong to the same label class), and we use the generated data in an unsupervised manner via MLM. In contrast, conventional data augmentation methods generate new utterances with the same label as the given utterance and utilize them in a supervised manner.


\subsection{Sequential Self-distillation}
To further boost performance, we employ self-distillation~\cite{mobahi2020self, allen2020towards} (Fig.~\ref{figure: idea_illustration}). The knowledge in the learned model is distilled into another model with the same architecture by matching their output logits\footnote{We have also tried to add a cross-entropy term~\cite{tian2020rethinking}, but find it hurts the performance.}:
\begin{equation}
  \theta_{k}= \argmin_{\theta_{k}} \text{KL}\left(\frac{\text{f}\left(\mathcal{D}; \theta_{k}\right)}{t}, \frac{\text{f}\left(\mathcal{D}; \theta_{k-1}\right)}{t}\right),
\label{equation: self-distillation KL}
\end{equation}
where $\text{KL}(\cdot)$ is the Kullback-Leibler (KL) divergence, $\text{f}(\cdot)$ is the output logit of the model, and $t$ is the temperature parameter. We adopt the born-again strategy~\cite{furlanello2018born} to iteratively distill the model into a sequence of generations. Hence, the model at $k_\text{th}$ generation with parameters $\theta_{k}$ is distilled to match the $(k-1)_\text{th}$ generation with parameters $\theta_{k-1}$.

\begin{table*}[t]
\begin{subtable}{1.0\textwidth}
\centering
\small
\begin{tabular}{lcccccccc}
\toprule
\multicolumn{1}{l}{\multirow{2}{*}{Method}} & 
\multicolumn{2}{c}{BANKING77} &
\multicolumn{2}{c}{HINT3} &
\multicolumn{2}{c}{HWU64} &
\multicolumn{2}{c}{MCID}  
\\
\cmidrule(lr){2-3}  \cmidrule(lr){4-5}  \cmidrule(lr){6-7} \cmidrule(lr){8-9}
& 5-shot & 10-shot & 5-shot & 10-shot & 5-shot & 10-shot & 5-shot & 10-shot\\
\midrule
TOD-BERT & 67.69\tiny(1.37) & 79.71\tiny(0.91) & 56.33\tiny(2.14)  & 66.42\tiny(2.19) & 74.83\tiny(1.11)  & 82.15\tiny(0.47)  & 66.37\tiny(2.65)  & 74.66\tiny(1.52)  \\
DNNC-NLI & 68.48\tiny(1.15) & 74.53\tiny(4.83) & 59.05\tiny(1.02) & 65.12\tiny(1.96) & 72.25\tiny(1.39)  & 77.91\tiny(1.11)  & 67.35\tiny(2.09)  & 75.20\tiny(1.28)  \\
DNNC-Intent & 70.36\tiny(1.85) & 78.85\tiny(1.56) & 58.08\tiny(4.98)  & 64.56\tiny(3.64) & 69.86\tiny(4.27)  & 74.87\tiny(3.02)  & 70.80\tiny(3.16)  & 78.60\tiny(1.49)  \\
CPFT & 70.96\tiny(2.45) & 79.44\tiny(.80) & 61.63\tiny(2.64) & 69.85\tiny(1.21) & 73.63\tiny(1.74) & 80.59\tiny(.61)   & 71.54\tiny(4.97) & 79.38\tiny(1.60)  \\
IntentBERT & 70.64\tiny(1.02) & 81.18\tiny(.34) & 58.96\tiny(1.50) &  68.96\tiny(1.50) & \textbf{77.60\tiny(.31)} & \textbf{83.55\tiny(.21)}   & 76.67\tiny(.84) & 81.60\tiny(1.41)   \\
IsoIntentBERT & 71.78\tiny(1.40) & \textbf{81.30\tiny(.50)} & 60.33\tiny(1.95) & 69.23\tiny(1.16) & \textbf{78.26\tiny(.69)}   & \textbf{83.70\tiny(.59)}  & \textbf{78.28\tiny(1.72)} & \textbf{82.51\tiny(1.23)}  \\
SE-Paraphrase & \textbf{71.92\tiny(.84)} & 81.18\tiny(.33) & \textbf{62.28\tiny(.77)} & \textbf{70.00\tiny(1.01)} & 76.75\tiny(.63)   & 82.88\tiny(.48)  & \textbf{78.32\tiny(2.12)} & \textbf{83.08\tiny(1.32)}  \\
SE-NLI & 70.03\tiny(1.47) & 80.58\tiny(1.13) & \textbf{61.69\tiny(1.59)} & 68.37\tiny(1.55) & 75.10\tiny(1.17)   & 82.57\tiny(.79)  & 74.54\tiny(1.86) & 81.20\tiny(1.80)  \\
\midrule
DFT & 69.01\tiny(1.54) & 78.92\tiny(1.69) &  60.65\tiny(1.60)  & 66.36\tiny(3.48) & 75.07\tiny(.53)  & 82.38\tiny(1.49)  & 72.32\tiny(1.80)  & 80.53\tiny(1.15)  \\
DFT++ (w/ CA) & \textbf{72.23\tiny{(1.80)}} & \textbf{82.33\tiny{(.72)}} & 60.53\tiny{(2.73)} & \textbf{70.36\tiny{(1.90)}} &  76.73\tiny{(1.05)} & 82.61\tiny{(.23)} & 77.45\tiny{(1.66)} & 81.27\tiny{(1.41)} \\
DFT++ (w/ SSD) & 68.86\tiny{(1.49)} & 80.32\tiny{(.81)} & 61.51\tiny{(1.88)} & 68.82\tiny{(2.49)} &  75.05\tiny{(1.36)} & 82.14\tiny{(.92)} & 74.17\tiny{(1.09)} & 81.44\tiny{(1.08)} \\
DFT++ (w/ CA, SSD) & \textbf{72.90\tiny{(.89)}} & \textbf{82.66\tiny{(.50)}} & \textbf{63.08\tiny{(1.17)}} & \textbf{70.47\tiny{(2.56)}} &  \textbf{77.73\tiny{(1.02)}} &  \textbf{83.45\tiny{(.38)}} & \textbf{79.43\tiny{(.84)}} & \textbf{82.83\tiny{(.76)}} \\
\bottomrule
\end{tabular}
\caption{BERT-based evaluation results.}
\label{table: main_result_bert_roberta: subtable, bert}
\end{subtable}
\newline
\begin{subtable}{1.0\textwidth}
\centering
\small
\begin{tabular}{lcccccccc}
\toprule
\multicolumn{1}{l}{\multirow{2}{*}{Method}} & 
\multicolumn{2}{c}{BANKING77} &
\multicolumn{2}{c}{HINT3} &
\multicolumn{2}{c}{HWU64} &
\multicolumn{2}{c}{MCID}  
\\
\cmidrule(lr){2-3}  \cmidrule(lr){4-5}  \cmidrule(lr){6-7} \cmidrule(lr){8-9}
& 5-shot & 10-shot & 5-shot & 10-shot & 5-shot & 10-shot & 5-shot & 10-shot\\
\midrule
DNNC-NLI & 73.90\tiny(1.27) & 79.51\tiny(2.56) & 59.73\tiny(0.89)  & 64.05\tiny(2.30) & 73.06\tiny(1.70)  & 78.12\tiny(1.86)  & 63.74\tiny(3.79)  & 73.72\tiny(1.82)  \\
DNNC-Intent & 72.97\tiny(1.46) & 77.69\tiny(5.06) & 61.15\tiny(1.74)  & 66.45\tiny(1.06) & 69.74\tiny(1.85)  & 72.30\tiny(3.61)  & 72.44\tiny(2.50)  & 78.64\tiny(1.69)  \\
CPFT & 70.94\tiny(1.08) & 78.57\tiny(.75) & 58.17\tiny(3.44) & 61.07\tiny(2.37) & 74.36\tiny(1.15) & 79.46\tiny(.81)   & 78.20\tiny(1.72) & 83.04\tiny(1.74)  \\
IntentRoBERTa & 75.23\tiny(.89) & 83.94\tiny(.33) & 60.77\tiny(1.60) &  68.91\tiny(1.24) & \textbf{78.97\tiny(1.26)} & 84.26\tiny(.84)   & 77.25\tiny(2.05) & 82.67\tiny(1.43)   \\
IsoIntentRoBERTa & 75.05\tiny(1.92) & 84.49\tiny(.43) & 59.79\tiny(2.72) & 69.08\tiny(1.59) & 78.09\tiny(1.06)   & 84.15\tiny(.58)  & 78.40\tiny(2.03) & 83.20\tiny(1.89)  \\
SE-Paraphrase & 76.03\tiny(.64) & 82.85\tiny(.89) & \textbf{63.96\tiny(.02)} & 69.14\tiny(2.08) & 76.50\tiny(.45)   & 81.25\tiny(.97)  & \textbf{80.78\tiny(1.36)} & 83.12\tiny(.86)  \\
SE-NLI & \textbf{76.56\tiny(.69)} & 84.65\tiny(.26) & 62.60\tiny(2.45) & 69.91\tiny(1.82) & 78.53\tiny(.84)   & 84.81\tiny(.45)  & \textbf{79.43\tiny(3.17)} & \textbf{84.13\tiny(1.25)}  \\
\midrule
DFT & 76.11\tiny(1.16) & 84.77\tiny(.43) & 61.39\tiny(1.51)  & 68.40\tiny(1.21) & 76.72\tiny(.94)  & 84.00\tiny(.34)  & 76.39\tiny(1.18)  & 82.55\tiny(1.15)  \\
DFT++ (w/ CA) & \textbf{78.74\tiny{(1.00)}} & \textbf{85.95\tiny{(.34)}} & \textbf{63.17\tiny{(2.20)}} & \textbf{71.30\tiny{(1.54)}} &  \textbf{79.02\tiny{(.89)}} &  \textbf{85.49\tiny{(.35)}} & 76.51\tiny{(2.77)} & \textbf{83.98\tiny{1.17)}} \\
DFT++ (w/ SSD) & 76.25\tiny{(1.67)} & \textbf{84.95\tiny{(.53)}} & 61.30\tiny{(2.31)} & \textbf{70.12\tiny{(1.35)}} &  77.57\tiny{(.62)} &  \textbf{84.91\tiny{(.45)}} & 78.73\tiny{(2.30)} & 83.37\tiny{1.64)} \\
DFT++ (w/ CA, SSD) & \textbf{78.90\tiny{(.50)}} & \textbf{86.14\tiny{(.19)}} & \textbf{63.61\tiny{(1.80)}} & \textbf{71.80\tiny{(1.88)}} & \textbf{79.93\tiny{(.92)}} &  \textbf{86.21\tiny{(.28)}} & \textbf{80.16\tiny{(2.74)}} & \textbf{84.80\tiny{(.79)}} \\
\bottomrule
\end{tabular}
\caption{RoBERTa-based evaluation.}
\label{table: main_result_bert_roberta: subtable, roberta}
\end{subtable}
\caption{Results of DFT++ and state-of-the-art methods. The mean value and standard deviation are reported. CA denotes context augmentation. SSD denotes sequential self-distillation. The top 3 results are highlighted.}
\label{table: main_result_bert_roberta}
\end{table*}
\begin{table}[h]
\centering
\small
\begin{tabular}{lcccc}
\toprule
 5-shot & Bank & Home  \\
\midrule
CINS$\ssymbol{5}$   & 89.1  &  80.2  \\
\cdashline{1-3}
DFT++ (BERT)   & \textbf{91.39\tiny(.78)}     & \textbf{82.11\tiny(4.09)}     \\
DFT++ (RoBERTa)   & \textbf{93.76\tiny(.46)}     & \textbf{86.21\tiny(2.94)}    \\
\midrule
 5-shot & Utility & Auto \\
\midrule
CINS$\ssymbol{5}$     & 95.4 & \textbf{93.7}  \\
\cdashline{1-3}
DFT++ (BERT)    & \textbf{96.16\tiny(.41)}    & 90.64\tiny(.93)     \\
DFT++ (RoBERTa)     & \textbf{97.39\tiny(.50)}    & \textbf{93.31\tiny(1.21)}     \\
\bottomrule
\end{tabular}
\caption{Comparison of DFT++ against CINS. $\ssymbol{5}$ denotes results copied from \citet{mi2022cins}. DFT++ is better in most cases, especially when RoBERTa is employed. The top 2 results are highlighted.}
\label{main_result_other_baseline}
\end{table}
Self-distillation can provably improve model performance if the data has a multi-view structure, i.e., the data has multiple features (views) to help identify its class~\cite{allen2020towards}. Such structures naturally exist in utterances. For instance, given the following utterance of label ``travel alert'',
\begin{itemize}
    \item[] \emph{``How safe is visiting Canada this week'',}
\end{itemize}
both ``safe'' and ``visiting'' indicate the intent label, 
and it is likely that the model learns only one of them because a single feature may be sufficient to discriminate the above utterance from others with different labels, especially with limited training data. Sequential self-distillation can help to learn both features, as shown in \citet{allen2020towards}.
\begin{figure}[h]
\centering
    \begin{subfigure}[b]{0.45\textwidth}
        \centering
        \includegraphics[scale=0.38]{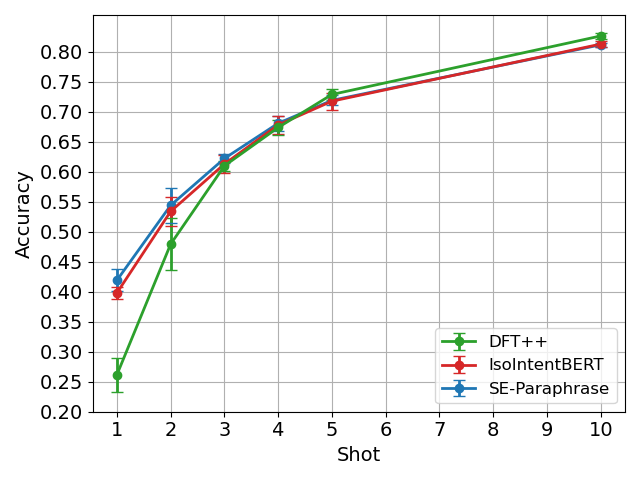}
        \caption{BERT-based experiments.}
        \label{figure: main_result_acc_vs_shot_bert}
    \end{subfigure}
     \\
    \begin{subfigure}[b]{0.45\textwidth}
        \centering
        \includegraphics[scale=0.38]{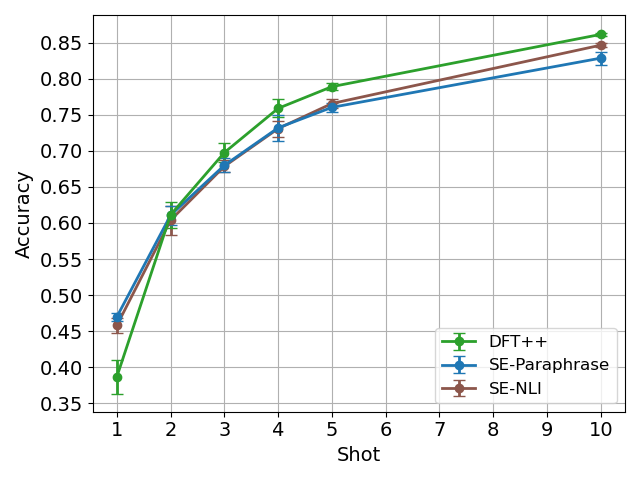}
        \caption{RoBERTa-based experiments.}
        \label{figure: main_result_acc_vs_shot_roberta}
    \end{subfigure}    
\caption{Impact of the size of labeled data on performance. The experiments are conducted on BANKING77. We compare DFT++ with the top 2 baselines.}
\label{figure: main_result_acc_vs_shot}
\end{figure}

\subsection{Experiments}
We evaluate DFT++ on the same benchmarks used to evaluate DFT. We compare DFT++ with state-of-the-art continual pre-training methods. Since early stopping is not necessary, as demonstrated in subsection~\ref{sec:DFT-experiments}, we combine the validation and test sets for a more comprehensive evaluation.


\paragraph{Baselines}
We compare with the following baselines. \textbf{TOD-BERT}~\cite{wu2020tod} conducts continual pre-training on dialogue corpus with MLM and response objectives. \textbf{DNNC-NLI}~\cite{zhang-etal-2020-discriminative} and \textbf{SE-NLI}~\cite{ma-etal-2022-effectiveness} employ NLI datasets. DNNC-NLI is equipped with a BERT-style pair-wise similarity model and a nearest neighbor classifier. SE-NLI employs sentence encoder~\cite{reimers2019sentence} with siamese and triplet architecture to learn the semantic similarity.
\textbf{DNNC-Intent}, \textbf{CPFT}~\cite{zhang-etal-2021-shot}, \textbf{IntentBERT}~\cite{zhang-etal-2021-effectiveness-pre} and \textbf{IsoIntentBERT}~\cite{zhang-etal-2022-fine} use external intent detection datasets. DNNC-Intent shares the same model structure as DNNC-NLI. CPFT adopts contrastive learning and MLM. IntentBERT employs standard supervised pre-training, based on which IsoIntentBERT introduces isotropization to improve performance. \textbf{SE-Paraphrase}~\cite{ma-etal-2022-effectiveness} exploits paraphrase corpus, using the same model architecture for sentence encode as SE-NLI. 

For all the baselines, we download the publicly released model if available. Otherwise, we follow the original work's guidelines to perform continual pre-training.
Next, we perform standard fine-tuning similar to DFT, using hyperparameters searched within the same range as our method, with three exceptions: DNNC-NLI, DNNC-Intent, and CPFT. For these methods, we use the original design and training configuration for in-task fine-tuning. 



In addition, we compare DFT++ against \textbf{CINS}~\cite{mi2022cins}, the most recent prompt-based method. CINS addresses intent detection by converting it into a cloze-filling problem through a carefully designed prompt template. Similar to our method, CINS directly fine-tunes PLMs on a limited amount of data.

\paragraph{Our method} We evaluate our method and the baselines based on two popular PLMs: BERT~\cite{devlin2018bert} and RoBERTa~\cite{liu2019roberta}. The representation of the token [CLS] is used as the utterance embedding. 
For a fair comparison, we select the hyper-parameters with the same validation data as used by the baselines, i.e., we follow IsoIntentBERT to use a portion of the OOS dataset as the validation data. The best hyper-parameters and the parameter range are given in the appendix.

\begin{table*}[t]
\begin{subtable}{1.0\textwidth}
\centering
\small
\begin{tabular}{lcccccccc}
\toprule
\multicolumn{1}{l}{\multirow{2}{*}{Method}} & 
\multicolumn{2}{c}{BANKING77} &
\multicolumn{2}{c}{HINT3} &
\multicolumn{2}{c}{HWU64} &
\multicolumn{2}{c}{MCID}  
\\
\cmidrule(lr){2-3}  \cmidrule(lr){4-5}  \cmidrule(lr){6-7} \cmidrule(lr){8-9}
& 5-shot & 10-shot & 5-shot & 10-shot & 5-shot & 10-shot & 5-shot & 10-shot\\
\midrule
DFT & 69.01\tiny(1.54) & 78.92\tiny(1.69) &  60.65\tiny(1.60)  & 66.36\tiny(3.48) & 75.07\tiny(.53)  & 82.38\tiny(1.49)  & 72.32\tiny(1.80)  & 80.53\tiny(1.15)  \\
\midrule
EDA & 68.81\tiny(1.97) & 72.97\tiny(.94) & 60.50\tiny(3.06) & 59.94\tiny(1.10) & 74.68\tiny(.81)  & 72.76\tiny(5.16)  & 73.10\tiny(.64)  & 80.99\tiny(.16)  \\
BT & 69.65\tiny(1.39) & 78.42\tiny(.83) & 60.50\tiny(1.40)  & 66.33\tiny(2.69) & 74.15\tiny(.84)  & 79.12\tiny(1.65)  & 75.15\tiny(2.04)  & 81.36\tiny(1.6)  \\
PrompDA & 71.62\tiny(.72) & 80.61\tiny(2.95) & \textbf{61.51\tiny(2.20)}  & 69.17\tiny(1.91) & 76.59\tiny(.89)  & \textbf{83.29\tiny(.56)}  & 77.16\tiny(.98)  & \textbf{81.47\tiny(2.19)}  \\
SuperGen &  64.83\tiny(1.06) & 77.48\tiny(0.37) & 57.30\tiny(1.41)  & 64.44\tiny(2.64) & 69.52\tiny(0.56)  & 77.26\tiny(0.88)  & 72.55\tiny(1.37)  & 78.78\tiny(1.01)  \\
GPT-J-DA & 71.84\tiny(1.41) & 78.34\tiny(.87) & 60.24\tiny(.83) & 67.40\tiny(2.41) & 70.72\tiny(.78) & 76.66\tiny(1.3)   & 73.92\tiny(2.77) & 78.77\tiny(2.39)  \\
\midrule
CA & \textbf{72.23\tiny{(1.80)}} & \textbf{82.33\tiny{(.72)}} & 60.53\tiny{(2.73)} & \textbf{70.36\tiny{(1.90)}} &  \textbf{76.73\tiny{(1.05)}} & 
82.61\tiny{(.23)} & \textbf{77.45\tiny{(1.66)}} & 81.27\tiny{(1.41)} \\
\bottomrule
\end{tabular}
\caption{BERT-based evaluation results.}
\label{table: table: analysis_context_aug_versus_data_aug_full: subtable, bert}
\end{subtable}
\newline
\begin{subtable}{1.0\textwidth}
\centering
\small
\begin{tabular}{lcccccccc}
\toprule
\multicolumn{1}{l}{\multirow{2}{*}{Method}} & 
\multicolumn{2}{c}{BANKING77} &
\multicolumn{2}{c}{HINT3} &
\multicolumn{2}{c}{HWU64} &
\multicolumn{2}{c}{MCID}  
\\
\cmidrule(lr){2-3}  \cmidrule(lr){4-5}  \cmidrule(lr){6-7} \cmidrule(lr){8-9}
& 5-shot & 10-shot & 5-shot & 10-shot & 5-shot & 10-shot & 5-shot & 10-shot\\
\midrule
DFT & 76.11\tiny(1.16) & 84.77\tiny(.43) & 61.39\tiny(1.51)  & 68.40\tiny(1.21) & 76.72\tiny(.94)  & 84.00\tiny(.34)  & 76.39\tiny(1.18)  & 82.55\tiny(1.15)  \\
\midrule
EDA & 74.74\tiny(1.08) & 81.84\tiny(.59) & 62.04\tiny(2.49) & 66.78\tiny(1.53) & 75.88\tiny(1.59)  & 81.91\tiny(.67)  & 77.17\tiny(1.85)  & 83.12\tiny(1.30)  \\
BT & 75.12\tiny(1.03) & 84.12\tiny(.28) & 60.83\tiny(1.16)  & 68.34\tiny(1.33) & 77.31\tiny(.72)  & 82.89\tiny(.21)  & 77.49\tiny(2.71)  & 82.05\tiny(1.45)  \\
PrompDA & 76.56\tiny(1.15) & 82.69\tiny(.99) & 60.56\tiny(1.37)  & 69.44\tiny(1.57) & 77.57\tiny(1.12)  & 82.94\tiny(1.29)  & 77.60\tiny(1.94)  & 83.86\tiny(2.27)  \\
SuperGen & 70.42\tiny(0.19) & 81.74\tiny(0.16) & 57.64\tiny(1.33)  & 65.88\tiny(0.54) &  71.28\tiny(0.78)  & 81.16\tiny(0.35)  & 73.99\tiny(1.79)  & 80.08\tiny(0.89)  \\
GPT-J-DA & 76.58\tiny(1.30) & 83.01\tiny(.87) & 62.16\tiny(1.83) & \textbf{71.45\tiny(1.86)} & 76.59\tiny(.94) & 81.65\tiny(.73)   & \textbf{77.91\tiny(2.22)} & 82.51\tiny(1.90)  \\
\midrule
CA & \textbf{78.74\tiny{(1.00)}} & \textbf{85.95\tiny{(.34)}} & \textbf{63.17\tiny{(2.20)}} & 71.30\tiny{(1.54)} &  \textbf{79.02\tiny{(.89)}} &  \textbf{85.49\tiny{(.35)}} & 76.51\tiny{(2.77)} & \textbf{83.98\tiny{1.17)}} \\
\bottomrule
\end{tabular}
\caption{Roberta-based evaluation results.}
\label{table: table: analysis_context_aug_versus_data_aug_full: subtable, roberta}
\end{subtable}
\caption{Comparison of our proposed contextual augmentation against conventional data augmentation methods. CA denotes contextual augmentation. The best results are highlighted.}
\label{table: analysis_context_aug_versus_data_aug_full}
\end{table*}
\paragraph{Main results}
We first examine the performance using a moderately small amount of data, specifically 5-shot and 10-shot scenarios. The results are summarized in Table~\ref{table: main_result_bert_roberta}. 
Remarkably, DFT++ performs comparably to a diverse set of baselines that leverage external resources, despite the fact that it solely utilizes the limited few-shot data available. The superiority of DFT++ can be attributed to the effective utilization of context augmentation and sequential self-distillation, both of which demonstrate improved results when applied independently in most cases. We notice that DFT++ performs better when using the stronger base model RoBERTa. As shown in Table~\ref{table: main_result_bert_roberta: subtable, roberta}, DFT++ outperforms all the baselines in most cases. Moreover, as shown in Table~\ref{main_result_other_baseline}, in most cases, DFT++ also outperforms CINS, the most recent prompt-based method, despite that CINS employs {T5-base}~\cite{raffel2020exploring} with 220 million parameters, which is almost twice the size of our base model.


To study the impact of the number of labeled data on performance, we reduce the number to only $1$ sample per label and present the results in Fig.~\ref{figure: main_result_acc_vs_shot}. We experiment with BANKING77, a challenging fine-grained dataset. When using BERT, we observe that DFT++ begins to outperform the baselines at a crossing point of $4$. When using RoBERTa, the crossing point is even smaller, at $2$, which is quite surprising. We have also observed similar phenomena on other datasets, as detailed in the appendix. The observations confirm our claim that the overfitting issue in directly fine-tuning PLMs for few-shot intent detection may not be as severe as initially presumed. The performance disadvantage resulting from overfitting can be effectively alleviated by leveraging other techniques to exploit the limited available data, even without resorting to the continual pre-training approach. 
However, in scenarios with an extremely small number of labeled data, the transferred knowledge from continual pre-training still provides significantly better performance compared to DFT++.

\subsection{Analysis}
\label{subsection: analysis}

\paragraph{Comparison between contextual augmentation and conventional data augmentation methods}
We compare our proposed context augmentation with the following conventional data augmentation methods. Easy Data Augmentation (EDA)~\cite{wei-zou-2019-eda} modifies a small number of utterances, e.g., through word swapping, to generate new augmented instances.
Back-translation (BT)~\cite{edunov-etal-2018-understanding} translates an utterance into another language and then translates it back\footnote{We use French as the intermediate language, and utilize {T5-base}~\cite{raffel2020exploring} and {opus-mt-fr-en}~\cite{TiedemannThottingal:EAMT2020} for translation.}. 
PromDA~\cite{wang-etal-2022-promda} and SuperGen~\cite{meng2022generating} are recent data augmentation methods leveraging generative PLMs.
GPT-J-DA~\cite{sahu-etal-2022-data} exploits the data generated by GPT-J in a supervised manner.
The results in Table~\ref{table: analysis_context_aug_versus_data_aug_full} show context augmentation is more robust against data shift. Note that SuperGen is designed for coarse-grained tasks with only two or three labels, such as sentiment classification. As a result, it may not scale effectively to intent detection tasks that involve a larger number of intents, typically ranging in the tens.
The comparison between context augmentation and GPT-J-DA highlights the superiority of unsupervised exploitation of the generated data. The inconsistent effectiveness of GPT-J-DA is also reported by ~\citet{sahu-etal-2022-data}.

\paragraph{Quality of context augmentation}
\label{analysis_augmentation_cases}
\begin{table*}[t]
\centering
\scriptsize
\begin{tabular}{p{5.0cm}p{5.0cm}p{5.0cm}}
\toprule
Input & Good & Bad \\
\midrule
``Is there a reason why my \textcolor{officegreen}{card was declined} when I attempted to \textcolor{officegreen}{withdraw money}?'', ``How come I can not \textcolor{officegreen}{get money} at the \textcolor{officegreen}{ATM}?'', ``Why can not I \textcolor{officegreen}{withdraw cash} from this \textcolor{officegreen}{ATM}?'', ``Why will not the \textcolor{officegreen}{ATM} give me \textcolor{officegreen}{cash}?'', ``This morning, I wanted to \textcolor{officegreen}{make a withdrawal} before work but my \textcolor{officegreen}{card was declined}, please double check it for me as this is the first time it \textcolor{officegreen}{was declined}.''
&
``\textcolor{officegreen}{ATM} will not let me \textcolor{officegreen}{withdraw my money} my card as \textcolor{officegreen}{refused} please help'', ``I \textcolor{officegreen}{withdrew} less than I expected from the \textcolor{officegreen}{ATM} on monday'', ``My \underline{\textcolor{officegreen}{wallet was stolen}} but my \textcolor{officegreen}{ATM card} was within safely'', ``I \underline{\textcolor{officegreen}{spent a fortune}} last week and have none left on my \textcolor{officegreen}{card} can you reverse \textcolor{officegreen}{refund the fees}'',``\underline{\textcolor{officegreen}{Please give me the code}} that I can use in the \textcolor{officegreen}{ATM} for my face to \textcolor{officegreen}{use my card}''
& 
``Why did my card never get a their villages and journey?'', ``An autofill took place but there was nothing to approve.'',  ``Can I get one form my card after I have made a ctifre?'', ``Family needs money for the holidays they said they can not make it I hope you can help even if it is not much.'' \\
\midrule
``Please \textcolor{officegreen}{order take} from Jasons Deli.'', ``Can you please \textcolor{officegreen}{order some food} for me?'', ``Can you look up Chinese \textcolor{officegreen}{takeout} near here?'', ``Can i \textcolor{officegreen}{order takeaway} from Spanish place?'', ``Find and \textcolor{officegreen}{order} rasgulla of janta sweet home pvt ltd.''
&
``I need to \textcolor{officegreen}{get some gluten free cookies} for my daughter'', ``Can you do ticket counter \textcolor{officegreen}{take away}'', ``How can I \textcolor{officegreen}{order Chinese food}'', ``\underline{\textcolor{officegreen}{Delivery service}} please \textcolor{officegreen}{order some takeaway} jahdi'', ``\textcolor{officegreen}{Order} beef kasundi bewa rasgulla and dosa \underline{\textcolor{officegreen}{will be ready}} in 10 mins''
&
``Please make some reservation if you want booking on myhotelcom'', ``Drive take from a taxi'', ``Warehouse 26723'', ``Please make some reservation if you want booking on  myhotelcom''
\\
\bottomrule
\end{tabular}
\caption{Utterances generated by GPT-J. The first row corresponds to the label ``Declined Cash Withdrawal'' from BANKING77. The second row corresponds to the label ``Takeaway Order'' from HWU64. Good examples exhibit semantic relevance to the input data, while bad examples are irrelevant. \textcolor{officegreen}{Green} words are highlighted to indicate semantic relevance, while the \underline{\textcolor{officegreen}{underlined}} words deviating the sentence from the original label.}
\label{table: analysis_examples_GPT3}
\end{table*}
To demonstrate the quality of the data generated by context augmentation,
we provide some good and bad examples of generated utterances in Table~\ref{table: analysis_examples_GPT3}. It is observed that GPT-J is able to generate grammatically fluent utterances that exhibit a high level of contextual relevance to the input utterances, which are utilized by DFT++ to better model the target data distribution. On the other hand, as also observed in~\citet{sahu-etal-2022-data}, 
some of the generated utterances deviate from the original label and, therefore, are not suitable for data augmentation. However, DFT++ mitigates this issue by focusing solely on leveraging contextual relevance, resulting in improved robustness against data shift (Table~\ref{table: analysis_context_aug_versus_data_aug_full}).

\paragraph{Complementarity of continual pre-training and DFT++}
\label{analysis_transfer_DFT++_complementarity}
Continual pre-training and DFT++ mitigate overfitting from different aspects. The former leverages external data, while the latter 
maximizes the utilization of the limited available data. Hence, it is likely that they are complementary. To support this claim, we present empirical results demonstrating their complementarity in Table~\ref{table: analysis, complementarity}. It is observed that when combined with DFT++, the two competitive methods, IsoIntentBERT and SE-Paraphrase, both demonstrate improved performance.

\begin{table}[h]
\renewrobustcmd{\bfseries}{\fontseries{b}\selectfont}
\renewrobustcmd{\boldmath}{}
\centering
\small
\begin{tabular}{wc{1.9cm}wc{1.3cm}wc{1.2cm}wc{1.2cm}}
\toprule
IsoIntentBERT & DFT++ & BANKING77 & HWU64 \\
\midrule
\checkmark &  & 71.78\tiny(1.40) & 78.26\tiny(.69) \\
\checkmark & \checkmark & \textbf{73.53\tiny(1.33)} & \textbf{80.20\tiny(1.20)} \\
\toprule
SE-Paraphrase & DFT++ & BANKING77 & HWU64 \\
\midrule
\checkmark &  & 71.92\tiny(.84) & 76.75\tiny(.63)\\
\checkmark & \checkmark  & \textbf{73.21\tiny(1.24)} & \textbf{78.34\tiny(.31)} \\
\bottomrule
\end{tabular}
\caption{Complementarity of DFT++ and continued pre-training with experiments conducted on 5-shot tasks.
}
\label{table: analysis, complementarity}
\end{table}

\paragraph{Impact of hyper-parameters} We study the impact of two key hyper-parameters, the size of the generated data and the number of self-distillation generations. As visualized in Fig.~\ref{figure: hyper-parameters, key parameters, subfigure, contextual augmented context size}, a positive correlation is found between the performance and the size of the augmented data.
The performance saturates after the data size per label reaches $50$. It is noted that when only the given data are used for MLM, i.e., the generated data size is $0$, MLM has an adversarial effect probably due to overfitting on the few given data. Such negative effect is successfully alleviated by context augmentation.  
As for self-distillation generations, 
we find that multiple generations of self-distillation are necessary to achieve better performance. 
In the appendix, we further analyze the impact of of the temperature parameters of GPT-J and self-distillation.
\begin{figure}[t]
\centering
    \begin{subfigure}[t]{0.23\textwidth}
        \centering
        \includegraphics[scale=0.24]{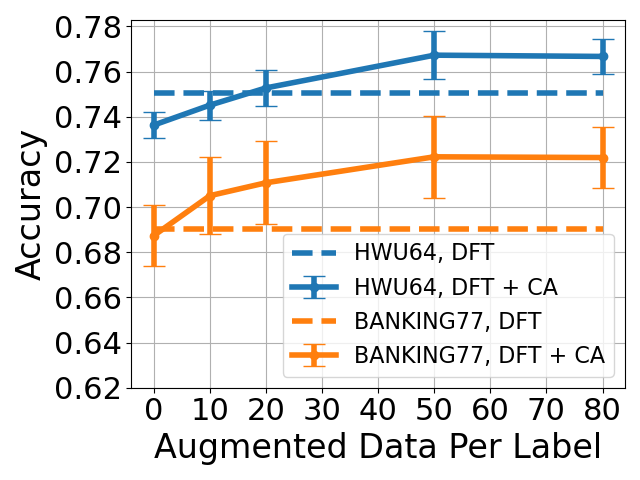}
        \caption{
        }
        \label{figure: hyper-parameters, key parameters, subfigure, contextual augmented context size}
    \end{subfigure}
    \begin{subfigure}[t]{0.23\textwidth}
        \centering
        \includegraphics[scale=0.24]{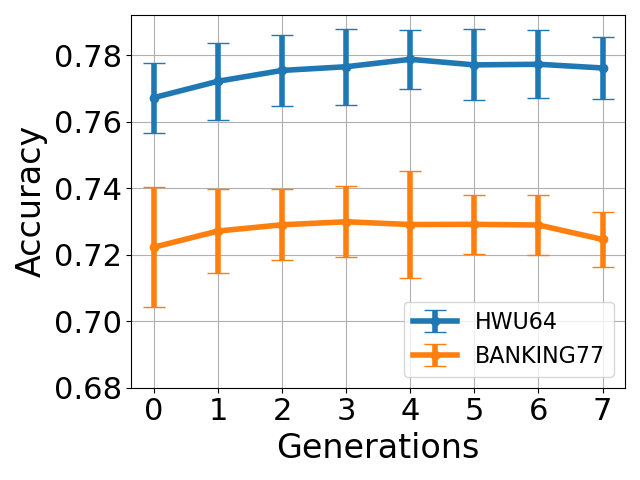}
        \caption{
        }
        \label{figure: hyper-parameters, key parameters, subfigure, self-distillation, generations}
    \end{subfigure}
\caption{Impact of the size of the augmented data (a) and the number of self-distillation generations (b). The experiments are conducted in $5$-shot scenarios. CA denotes context augmentation.
}
\label{figure: hyper-parameters, key parameters}
\end{figure}

\paragraph{Comparison with alternative context augmentation methods}
\begin{table}[h]
\centering
\small
\begin{tabular}{lcc}
\toprule
\multicolumn{1}{l}{\multirow{2}{*}{Method}} & \multicolumn{2}{c}{BANKING77}  \\
\cmidrule(lr){2-3}
& 5-shot & 10-shot \\
\midrule
DFT & 69.01\tiny(1.54) & 78.92\tiny(1.69) \\
DFT + External   & 67.84\tiny(.82) & 81.23\tiny(.66) \\ 
DFT + EDA   & 70.61\tiny(1.78) & 81.83\tiny(.41) \\ 
\midrule
DFT + GPT-J   & \textbf{72.22\tiny(1.80)} & \textbf{82.33\tiny(.72)} \\ 
\bottomrule
\end{tabular}
\caption{Comparison of our proposed GPT-J-based context augmentation with other alternatives. ``External'' denotes a corpus collected from Wikipedia.}
\label{analysis_alternative_context_augmentation}
\end{table}

We have also studied alternative context augmentation methods. The first one is Easy Data Augmentation~(EDA)~\cite{wei-zou-2019-eda} with random synonym replacement, insertion, swap, and deletion. 
The second approach involves manually collecting a domain-specific corpus. We conduct experiments on BANKING77, since it focuses on a single domain, making it convenient to collect the corpus. We extract web pages from Wikipedia\footnote{https://en.wikipedia.org} with keywords that are closely relevant to ``Banking'', such as ``Bank'' and ``Credit card''. The keywords can be found in the appendix. As shown by Table~\ref{analysis_alternative_context_augmentation}, our GPT-J-based context augmentation outperforms the alternatives. We attribute the superiority to the grammatical fluency achieved by leveraging the generative power of GPT-J, which is typically compromised by EDA. Additionally, the high degree of semantic relevance observed in our approach is rarely guaranteed in the noisy corpus collected from Wikipedia.
\section{Related Works}
\paragraph{Few-shot Intent Detection} Before the era of PLMs, the study of few-shot intent detection focuses on model architecture~\cite{geng2019few, xia2020composed, nguyen2020dynamic}. Recently, fine-tuning PLMs has become the mainstream methodology. \citet{zhang-etal-2020-discriminative} fine-tune pair-wise encoder on natural language inference~(NLI) tasks. \citet{zhang-etal-2021-shot} fine-tune PLMs in a contrastive manner. \citet{zhang-etal-2021-effectiveness-pre} leverage public intent detection dataset, which is further improved by isotropization~\cite{zhang-etal-2022-fine}. Other settings are also studied, including semi-supervised learning~\cite{dopierre2020few, dopierre-etal-2021-protaugment} and incremental learning~\cite{xia-etal-2021-incremental}. Unlike the mainstream strategy, our method does not require continual pre-training on extra resources.

\paragraph{Continual Pre-training of PLMs} 
Continual pre-training of PLMs is helpful~\cite{gururangan2020don, ye-etal-2021-crossfit, luo2021bi}. For dialogue understanding, many works leverage conversational corpus to perform continual pre-training. \citet{cp-dialogue1} conducts continual pre-training with a dialogue-adaptive pre-training objective and a synthesized in-domain corpus. \citet{cp-tod-bert} further pre-trains BERT with dialogue corpora through masked language modeling and contrastive loss. \citet{cp-convert-bert} use Reddit conversational corpus to pre-train a dual-encoder model. \citet{vulic-etal-2021-convfit} adopts adaptive conversational fine-tuning on a dialogue corpus.


\paragraph{PLM-based Data Augmentation} \citet{rosenbaum-etal-2022-linguist} fine-tune 
PLMs to generate data for intent detection and slot tagging. \citet{jolly-etal-2020-data} develop novel sampling strategies to improve the generated utterances. \citet{kumar-etal-2022-controlled} pre-train a token insertion PLM for utterances generation. However, these methods require slot values, which are assumed unavailable in this work. \citet{papangelis-etal-2021-generative} fine-tune PLMs with reinforcement learning, but our augmentation method adopts off-the-shelf PLM without further training. The closest work to ours is~\citet{sahu-etal-2022-data}, which utilizes off-the-shelf PLMs for data augmentation. 
However, our method focuses solely on leveraging contextual relevance to achieve improved robustness. PLM-based data augmentation has been explored for other tasks, e.g. sentiment classification ~\cite{yoo-etal-2021-gpt3mix-leveraging, wang-etal-2022-promda, chen-liu-2022-rethinking} and natural language inference~\cite{meng2022generating, ye2022zerogen}. However, these approaches may fail to scale to intent detection tasks with tens of intent classes, as shown by ~\citet{sahu-etal-2022-data} and our experiments.

\section{Conclusions and Limitations}
We revisit few-shot intent detection with PLMs by comparing two approaches: direct fine-tuning and continual pre-training. 
We show that the overfitting issue may not be as significant as commonly believed. In most cases, our proposed framework, DFT++, demonstrates superior performance compared to mainstream continual pre-training methods that rely on external training corpora.


One limitation of DFT++ is the computational overhead caused by generative PLMs. 
Additionally, our current approach includes all utterances generated by the PLM, even those that might lack contextual relevance or contain noise. These issues are left for future exploration.

\section*{Acknowledgments}
We would like to thank the anonymous reviewers for their helpful comments. This research was partially supported by the grant of HK ITF ITS/359/21FP.


\newpage
\appendix
\section{Appendix}
\label{sec:appendix}


\paragraph{Hyper-parameters}
We determine the hyper-parameters by grid search. The best hyper-parameters and the search range are summarized in Table~\ref{table: hyper-parameters} and Table~\ref{table: hyper-parameters_range}, respectively. The grid search is performed with OOS dataset. Specifically, we follow IsoIntentBERT to use the two domains ``Travel'' and ``Kitchen dining'' as the validation set. To guarantee a fair comparison, the same validation set is also employed for all the baselines.

\begin{table}[h]
\centering
\small
\begin{tabular}{lp{5.5cm}}
\toprule
PLM   &  Hyper-parameter \\
\midrule
BERT      &  $\text{lr}_\text{PLM}=2e-4$, $\text{lr}_\text{cls}=2e-5$, $\lambda=1.0$, context\_size =50, $t=100$, $\text{iteration}$=6.  \\
\midrule
RoBERTa   &  $\text{lr}_\text{PLM}=2e-5$, $\text{lr}_\text{cls}=2e-3$, $\lambda=0.1$, context\_size =50, $t=40$,  $\text{iteration}$=5. \\
\bottomrule
\end{tabular}
\caption{Hyper-parameters of DFT++. $\text{lr}_\text{PLM}$ and $\text{lr}_\text{cls}$ denote the learning rate of the PLM and the linear classifier, respectively. context\_size is the size of the augmented contextual utterances per label. iteration is the number of iterations/generations in sequential self-distillation.}
\label{table: hyper-parameters}
\end{table}
\begin{table}[h]
\centering
\small
\begin{tabular}{lp{4.5cm}}
\toprule
Parameter &  Range \\
\midrule
$\text{lr}_\text{PLM}$ & $\{2e-5, 2e-4, 2e-3\}$ \\
$\text{lr}_\text{cls}$ & $\{2e-5, 2e-4, 2e-3\}$ \\
$\lambda$ & $\{0.01, 0.1, 1.0, 10.0\}$ \\
$\text{context\_size}$ & $\{1, 2, 5, 10, 20, 50, 80\}$ \\
$t$ & $\{0.1, 1, 10, 40, 80, 100, 200, 500\}$ \\
$\text{iteration}$  & $\{1,2,3,4,5,6,7\}$ \\
\bottomrule
\end{tabular}
\caption{Grid search range of hyper-parameters.}
\label{table: hyper-parameters_range}
\end{table}

\paragraph{Implementation details} We use Python, PyTorch library and Hugging Face library~\footnote{https://github.com/huggingface/transformers} to implement the model. We adopt \emph{bert-base-uncased} and \emph{roberta-base} with around $110$ million parameters. We use AdamW as the optimizer. We use different learning rates for PLMs and the linear classifier, determined by grid-search. The parameter for weight decay is set to $1e-3$. We employ a linear scheduler with the warm-up proportion of $5\%$. We fine-tune the model for $200$ epochs to guarantee convergence. The experiments are conducted with Nvidia RTX 3090 GPUs. We repeat all experiments for $5$ times, reporting the averaged accuracy and standard deviation. 

\begin{figure}[h]
\centering
    \begin{subfigure}[t]{0.23\textwidth}
        \centering
        \includegraphics[scale=0.24]{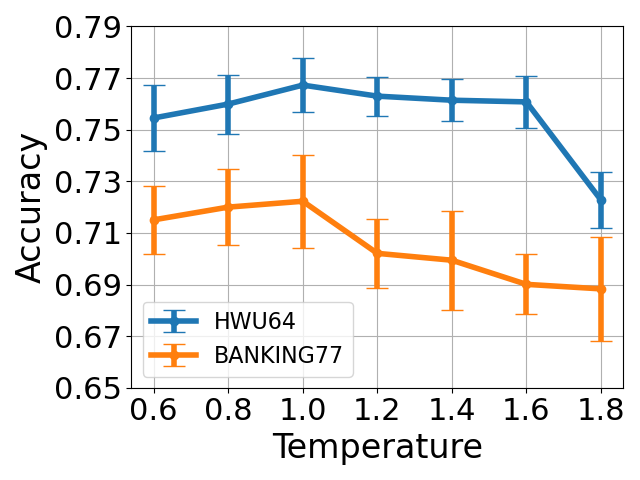}
        \caption{
        }
        \label{figure: hyper-parameters, two temperatures, subfigure, GPT-J temperature}
    \end{subfigure}
    \begin{subfigure}[t]{0.23\textwidth}
        \centering
        \includegraphics[scale=0.24]{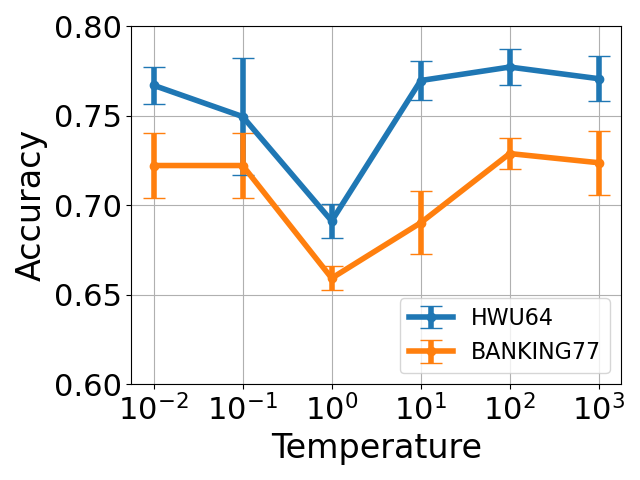}
        \caption{
        }
        \label{figure: hyper-parameters, two temperatures, subfigure, self-distillation, temperature}
    \end{subfigure}
\caption{Impact of the temperature parameter of GPT-J (a) and self-distillation (b). The experiments are conducted in $5$-shot scenarios.}
\label{figure: hyper-parameters, two temperatures}
\end{figure}

\begin{figure*}[h]
\centering
    \begin{subfigure}[b]{0.45\textwidth}
        \centering
        \includegraphics[scale=0.40]{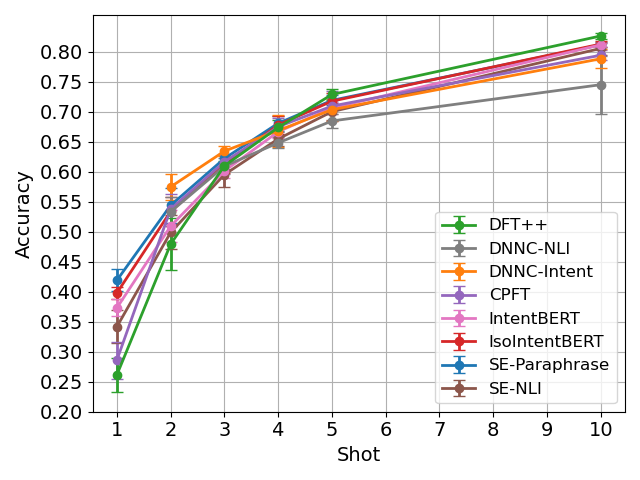}
        \caption{BERT-based experiments on BANKING77.}
        \label{figure: main_result_acc_vs_shot_bert_banking}
    \end{subfigure}
    \begin{subfigure}[b]{0.45\textwidth}
        \centering
        \includegraphics[scale=0.40]{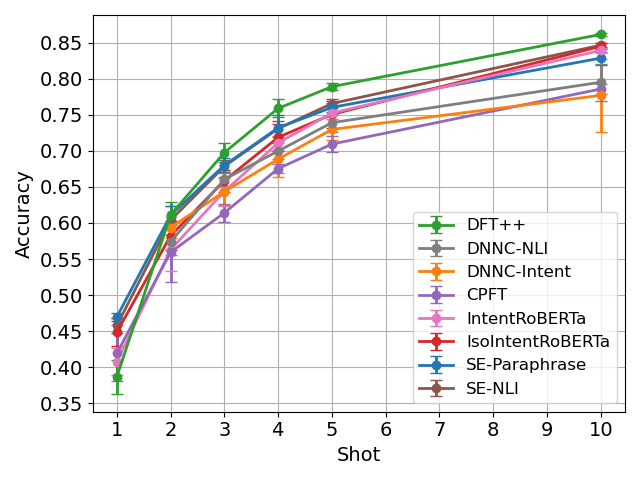}
        \caption{RoBERTa-based experiments on BANKING77.}
        \label{figure: main_result_acc_vs_shot_roberta_banking}
    \end{subfigure}    
    \begin{subfigure}[b]{0.45\textwidth}
        \centering
        \includegraphics[scale=0.40]{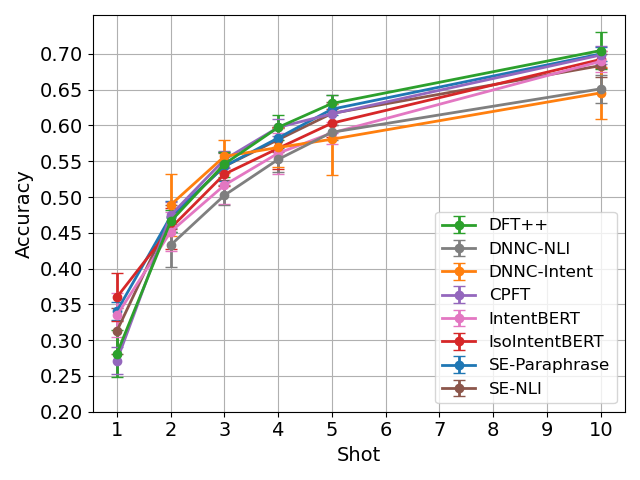}
        \caption{BERT-based experiments on HINT3.}
        \label{figure: main_result_acc_vs_shot_bert_hint3}
    \end{subfigure}
    \begin{subfigure}[b]{0.45\textwidth}
        \centering
        \includegraphics[scale=0.40]{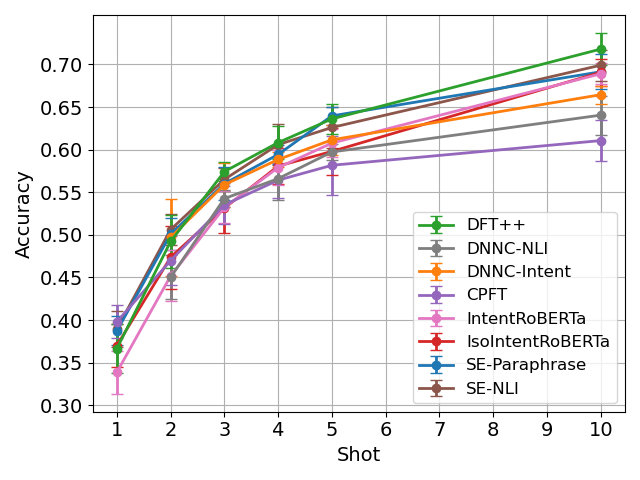}
        \caption{RoBERTa-based experiments on HINT3.}
        \label{figure: main_result_acc_vs_shot_roberta_hint3}
    \end{subfigure}    
    \begin{subfigure}[b]{0.45\textwidth}
        \centering
        \includegraphics[scale=0.40]{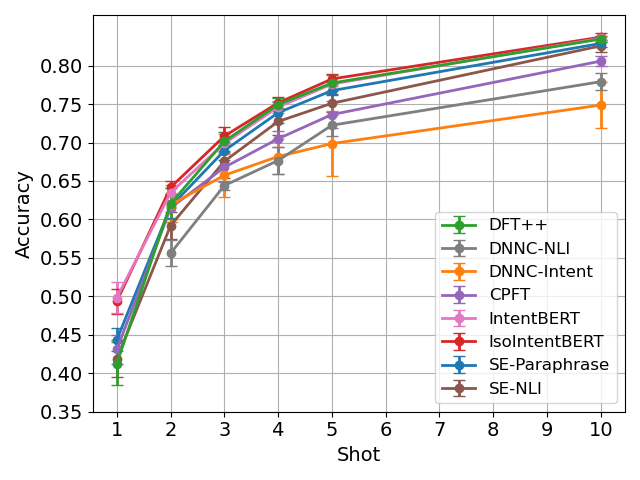}
        \caption{BERT-based experiments on HWU64.}
        \label{figure: main_result_acc_vs_shot_bert_hwu64}
    \end{subfigure}
    \begin{subfigure}[b]{0.45\textwidth}
        \centering
        \includegraphics[scale=0.40]{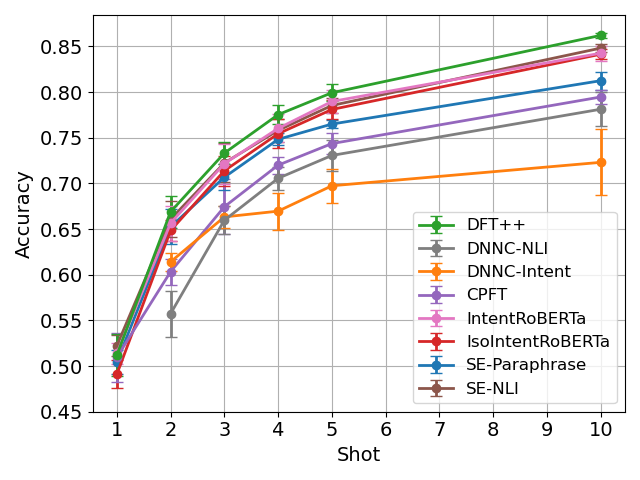}
        \caption{RoBERTa-based experiments on HWU64.}
        \label{figure: main_result_acc_vs_shot_roberta_hwu64}
    \end{subfigure}    
    \begin{subfigure}[b]{0.45\textwidth}
        \centering
        \includegraphics[scale=0.40]{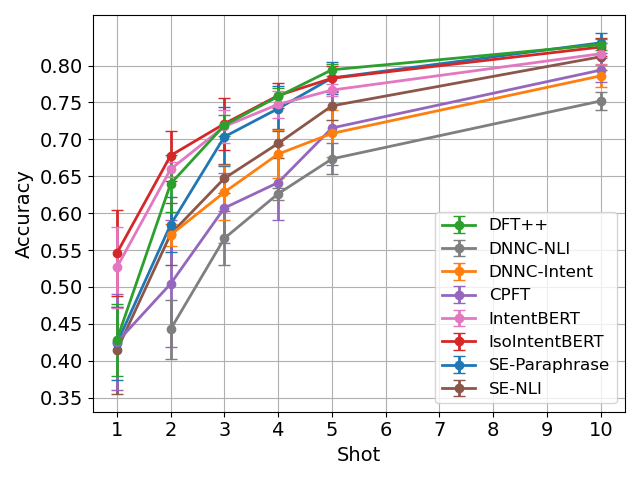}
        \caption{BERT-based experiments on MCID.}
        \label{figure: main_result_acc_vs_shot_bert_mcid}
    \end{subfigure}
    \begin{subfigure}[b]{0.45\textwidth}
        \centering
        \includegraphics[scale=0.40]{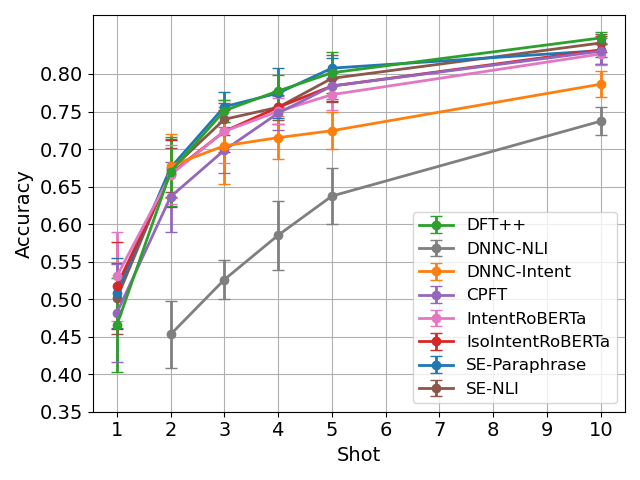}
        \caption{RoBERTa-based experiments on MCID}
        \label{figure: main_result_acc_vs_shot_roberta_mcid}
    \end{subfigure}    
\caption{Impact of the number of labeled data on model performance.}
\label{figure: main_result_acc_vs_shot_more_result}
\end{figure*}

\paragraph{Impact of the number of labeled data on performance}
We provide the full results in Fig.~\ref{figure: main_result_acc_vs_shot_more_result}. It is observed that DFT++ outperforms many competitive methods fine-tuned on extra data even when the number of labeled data is small.

\paragraph{Keywords used to collect the corpus for an alternative context augmentation method}
As introduced in subsection~\ref{subsection: analysis}, one alternative context augmentation method involves manually collecting a domain-specific corpus. We experiment with BANKING77. To collect an external corpus, we extract web pages from Wikipedia\footnote{https://en.wikipedia.org} with keywords closely related to ``Banking'', such as ``Bank'' and ``Credit card''. The adopted keywords are summarized in Table~\ref{table: wiki_key_words}.
\begin{table}[h]
\centering
\small
\begin{tabular}{p{0.40\textwidth}}
\toprule
``Bank'', ``Credit'', ``Debt'', ``Payment'', ``Fund'', ``Credit card'', ``Banking agent'', ``Bank regulation'', ``Cheque'', ``Coin'', ``Deposit account'', ``Electronic funds transfer'', ``Finance'', ``Internet banking'', ``Investment banking'', ``Money'', ``Wire transfer'', ``Central bank'', ``Credit union'', ``Public bank'', ``Cash'', ``Call report'', ``Ethical banking'', ``Loan'', ``Mobile banking'', ``Money laundering'', ``Narrow banking'', ``Private banking'' \\
\bottomrule
\end{tabular}
\caption{Key words used to collect the corpus from Wikipedia.}
\label{table: wiki_key_words}
\end{table}

\paragraph{Analysis of hyper-parameters}
We show the impact of the temperature parameter of GPT-J and self-distillation in Fig.~\ref{figure: hyper-parameters, two temperatures}. The temperature parameter of GPT-J controls the diversity of the generated context. A higher temperature makes the generated text more diverse. As shown in the figure, the best performance is reached when the diversity is moderate.For self-distillation, both small and large temperatures can produce good results.

\label{sec:appendix}



\end{document}